\documentclass[preprint]{elsarticle}
\usepackage[utf8]{inputenc}
\usepackage[hidelinks]{hyperref}
\usepackage{xcolor}
\hypersetup{
    colorlinks,
    linkcolor={red!50!black},
    citecolor={blue!50!black},
    urlcolor={blue!80!black}
}
\usepackage{amsmath}
\usepackage{amsfonts}
\usepackage{amssymb}
\usepackage{tikz,ifthen}
\usepackage{pgfplots}
\usepackage{subcaption}
\usepackage{booktabs}
\usepackage{needspace} % To ensure enough space for subsections
\usepackage{float}
\usepackage{siunitx}

\usepackage{tikz}
\usepackage{listofitems} % for \readlist to create arrays
\tikzstyle{mynode}=[thick,draw=blue,fill=blue!20,circle,minimum size=22]
\usetikzlibrary{positioning}
\usetikzlibrary{arrows.meta}

\usepackage{xcolor}
\colorlet{myred}{red!80!black}
\colorlet{myblue}{blue!80!black}
\colorlet{mygreen}{green!60!black}
\colorlet{myorange}{orange!70!red!60!black}
\colorlet{mydarkred}{red!30!black}
\colorlet{mydarkblue}{blue!40!black}
\colorlet{mydarkgreen}{green!30!black}
\tikzstyle{node}=[thick,circle,draw=myblue,minimum size=22,inner sep=0.5,outer sep=0.6]
\tikzstyle{node in}=[node,green!20!black,draw=mygreen!30!black,fill=mygreen!25]
\tikzstyle{node hidden}=[node,blue!20!black,draw=myblue!30!black,fill=myblue!20]
\tikzstyle{node convol}=[node,orange!20!black,draw=myorange!30!black,fill=myorange!20]
\tikzstyle{node out}=[node,red!20!black,draw=myred!30!black,fill=myred!20]
\tikzstyle{connect}=[thick,mydarkblue] %,line cap=round
\tikzstyle{connect arrow}=[-{latex[length=4,width=3.5]},thick,mydarkblue,shorten <=0.5,shorten >=1]
\tikzset{ % node styles, numbered for easy mapping with \nstyle
  node 1/.style={node in},
  node 2/.style={node hidden},
  node 3/.style={node out},
}
\def\nstyle{int(\lay<\Nnodlen?min(2,\lay):3)} % map layer number onto 1, 2, or 3

\usepackage{placeins} % To use the \FloatBarrier command
\usepackage[inline]{asymptote}

\usepackage{natbib}
\biboptions{super,comma,sort&compress}
\makeatletter
\renewcommand\@biblabel[1]{#1.}
\renewcommand\@cite[2]{#1\if@tempswa, #2\fi}
\makeatother

\newcommand{\defineproblem}[2]{%
  \hypertarget{#1}{#2}%
}

\newcommand{\problemref}[1]{%
  Problem~\hyperlink{#1}{#1}%
}

\newcommand{\problemreff}[1]{%
  \hyperlink{#1}{#1}%
}

\newcommand{\appropto}{\mathrel{\vcenter{
  \offinterlineskip\halign{\hfil$##$\cr
    \propto\cr\noalign{\kern2pt}\sim\cr\noalign{\kern-2pt}}}}}

\makeatletter
\let\@afterindenttrue\@afterindentfalse
\makeatother

\DeclareMathOperator{\atantwo}{atan2}

\makeatletter
\providecommand*{\diff}%
{\@ifnextchar^{\DIfF}{\DIfF^{}}}
\def\DIfF^#1{%
\mathop{\text{\mathstrut d}}%
\nolimits^{#1}\gobblespace}
\def\gobblespace{%
\futurelet\diffarg\opspace}
\def\opspace{%
\let\DiffSpace\!%
\ifx\diffarg(%
\let\DiffSpace\relax
\else
\ifx\diffarg[%
\let\DiffSpace\relax
\else
\ifx\diffarg\{%
\let\DiffSpace\relax
\fi\fi\fi\DiffSpace}

\NewDocumentCommand{\evalat}{sO{\big}mm}{%
  \IfBooleanTF{#1}
   {\mleft. #3 \mright|_{#4}}
   {#3#2|_{#4}}%
}
\newcommand{\norm}[1]{\left\lVert #1 \right\rVert}

\journal{Journal of Computational Mathematics and Data Science}

\newcommand{\params}{\boldsymbol{\theta}}
\newcommand{\mappingSol}{\mathbf{m}}
\newcommand{\mappingApprox}{\mathbf{m}_{\params}}
\newcommand{\pdeApprox}{u_{\params}}
\newcommand{\pdeSol}{u}
\newcommand{\sourceDist}{f}
\newcommand{\targetDist}{g}
\newcommand{\sourceDomain}{\mathcal{S}}
\newcommand{\targetDomain}{\mathcal{T}}

\newcommand{\problemcaption}[2]{%
  \let\oldthefigure\thefigure
  \renewcommand{\thefigure}{\arabic{figure} [\protect\problemref{#1}]}%
  \caption{#2}%
  \let\thefigure\oldthefigure
}

\pgfplotsset{compat=1.18} 
\begin{document}

\begin{frontmatter}

\title{A neural network approach for solving the Monge-Ampère equation with transport boundary condition}
\tnotetext[t1]{This work in the project MALIOD is funded by Holland High Tech \textbar\ TKI HSTM via the PPS allowance scheme for public-private partnerships. }

\author[inst1]{Roel Hacking\corref{cor1}}
\ead{r.g.j.hacking@tue.nl}

\author[inst1]{Lisa Kusch}
\author[inst1]{Koondanibha Mitra}
\author[inst1]{Martijn Anthonissen}
\author[inst1,inst2]{Wilbert IJzerman}

\affiliation[inst1]{organization={Eindhoven University of Technology},%Department and Organization
            % addressline={Address One}, 
            addressline={PO Box 513},
            postcode={5600 MB}, 
            city={Eindhoven},
            % state={North Brabant},
            country={The Netherlands}}

\affiliation[inst2]{organization={Signify},%Department and Organization
            % addressline={Address One}, 
            addressline={High Tech Campus 7},
            postcode={5656 AE}, 
            % state={North Brabant},
            city={Eindhoven},
            country={The Netherlands}}

\cortext[cor1]{Corresponding author}

\begin{abstract}
% This paper presents a novel approach to solving the Monge-Ampère equation with transport boundary condition using artificial neural networks, with applications to optical design. We formulate a loss function that incorporates the residual of the partial differential equation, boundary condition, and convexity constraints, allowing multilayer perceptron networks to be trained as approximate solutions. The method is evaluated on several test problems, including symmetric and asymmetric circle-to-circle mappings, as well as square-to-circle and circle-to-flower mappings. We compare our method's results with a more traditional least-squares finite-difference solver in terms of accuracy and convergence speed. Our neural network approach, optimized using L-BFGS, demonstrates competitive and often superior performance in all test cases. We also performed a hyperparameter study that examined the effects of sampling density, network architecture, and optimization algorithm. While showing promise, further research is necessary to establish the robustness of the method for more complex problems and to ensure consistent convergence. The simplicity and flexibility of this neural network-based method make it an attractive alternative to specialized partial differential equation solvers. 
This paper introduces a novel neural network-based approach to solving the Monge-Ampère equation with the transport boundary condition, specifically targeted towards optical design applications. We leverage multilayer perceptron networks to learn approximate solutions by minimizing a loss function that encompasses the equation's residual, boundary conditions, and convexity constraints. Our main results demonstrate the efficacy of this method, optimized using L-BFGS, through a series of test cases encompassing symmetric and asymmetric circle-to-circle, square-to-circle, and circle-to-flower reflector mapping problems. Comparative analysis with a conventional least-squares finite-difference solver reveals the competitive, and often superior, performance of our neural network approach on the test cases examined here. A comprehensive hyperparameter study further illuminates the impact of factors such as sampling density, network architecture, and optimization algorithm. While promising, further investigation is needed to verify the method's robustness for more complicated problems and to ensure consistent convergence. Nonetheless, the simplicity and adaptability of this neural network-based approach position it as a compelling alternative to specialized partial differential equation solvers.
\end{abstract}

\end{frontmatter}

\section{Introduction}
The Monge-Ampère equation is a nonlinear partial differential equation (PDE) with crucial applications across various fields in physics and mathematics. Its general form is given by:

\begin{equation} 
\det\left(D^2 u\right) = f(x, u, \nabla u), 
\end{equation}

\noindent where $u: \mathbb{R}^N \to \mathbb{R}$, ($N \geq 1$), is an unknown convex function, and $D^2 u$ represents the Hessian matrix of $u$. This equation traces its origins back to the 18th-century work of Gaspard Monge, who studied the problem of optimal resource allocation. Over time, this foundational problem has evolved into what is now known as the optimal transport problem, a concept that naturally emerges in fields such as fluid dynamics and mathematical physics. The Monge-Ampère equation effectively describes the optimal transportation of one mass distribution to another, minimizing a cost function that typically represents the distance over which each mass element must be moved \cite{de2014monge}.

Several optical problems can be formulated as instances of optimal transport. A notable example is the design of a reflector that transforms a given light source distribution into a desired target distribution, a problem that inherently aligns with the principles of optimal transport. In this context, the 'mass' represents energy, while the cost function corresponds to the optical path lengths of the light rays \cite{glimm2003optical}.

Numerous variations of this problem can be effectively modeled using the Monge-Ampère equation or its generalized forms. For example, the light source may be planar, emitting a parallel beam, or point-based, radiating in multiple directions. Additionally, optical systems can target near-field or far-field regions and may involve point or parallel targets. Both reflectors and lenses can be described within this framework. For example, we might wish to transform a point source to a far-field target using a freeform reflector; a numerical method for solving this problem using the intersection of confocal paraboloids has been described by \citet{de2016far}. For an in-depth exploration of these variations, see \citet{romijn2021generated} and \citet{Anthonissen:21}.

To simplify our analysis, we will focus on a specific optical configuration: the parallel-to-far-field reflector system. In this setup, a planar light source emits a parallel beam of light toward a reflector, and our primary concern is the distribution of the reflected light at a significant distance from the reflector. Consequently, we only need to consider the \textit{direction} of each reflected ray. By applying the Monge-Ampère equation with the transport boundary condition and solving it, we can determine the convex reflector surface that transforms a given source light distribution into the desired target distribution. It is important to note that, mathematically, this problem is identical to both the parallel-to-parallel reflector problem and the parallel-to-far-field lens problem, which can also be addressed using the method presented here.

This paper introduces a novel numerical method based on artificial neural networks (ANNs) to solve the Monge-Ampère equation with transport boundary condition. Numerous studies have explored the application of neural networks and automatic differentiation to solve ordinary and partial differential equations. \citet{dissanayake1994neural} pioneered this approach, demonstrating the potential of neural networks for approximating solutions to PDEs. Building on this work, \citet{712178} presented a method using ANNs to solve initial and boundary value problems by constructing trial solutions that inherently satisfy the given conditions. \citet{aarts2001neural} proposed a method to solve PDEs and their boundary/initial conditions using neural networks, incorporating knowledge about the PDE in the structure and training process of the network. More recently, \citet{michoski2020solving} presented a comprehensive study on the solving of differential equations using deep neural networks, demonstrating their competitiveness with conventional numerical methods and their potential for parameter space exploration and model enrichment. Building on this rich body of work, \citet{nystrom2023solving} employed ANNs to solve the Dirichlet problem for the Monge-Ampère equation. We extend this approach by incorporating the transport boundary condition and compare our neural network-based method against an existing numerical solver for the Monge-Ampère equation with transport boundary condition\cite{prins2015least}. Furthermore, we examine the effect of various hyperparameters of this neural network method on its performance.

Section \ref{sec:background} provides a concise background on the Monge-Ampère equation in the context of the parallel-to-far-field reflector problem, a previously proposed finite-differences-based solver for this problem, and artificial neural networks. In Section \ref{sec:methods}, we present our extension of this method to incorporate the alternative boundary conditions required for optimal transport problems. To demonstrate the effectiveness of our proposed method, Section \ref{sec:results} presents results for several example problems. As is common in machine learning, numerous hyperparameters influence the accuracy of our method. Thus, Section \ref{sec:results-hyper} empirically examines the effects of select hyperparameters on our method's performance. In Section \ref{sec:discussion} and Section \ref{sec:conclusion}, we conclude by discussing the advantages and limitations of neural network-based methods for solving the Monge-Ampère equation with the transport boundary condition, and explore potential avenues for future research to mitigate these limitations.

\section{Background}
\label{sec:background}
\subsection{Parallel-to-far-field reflector problem}
\label{sec:background-ma}
\begin{figure*}
\centering
\includegraphics[width=0.7\linewidth]{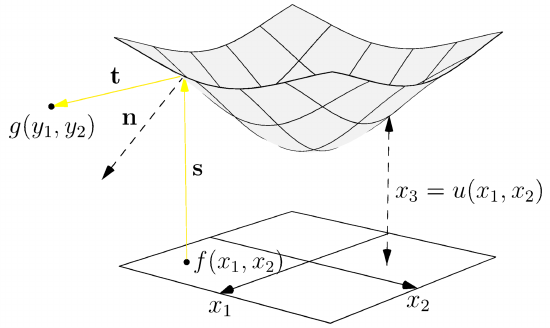}
\caption{Visualization of the parallel-to-far-field reflector problem}
\label{fig:reflector-system}
\end{figure*}
Consider a planar light source situated within three-dimensional Euclidean space $\mathbb{R}^3$ with Cartesian coordinates $(x_1, x_2, x_3)$, aligned parallel to the $x_1x_2$-plane (see Figure \ref{fig:reflector-system}), i.e. $x_3=0$. This light source emits light in the direction $\mathbf{s} = (0, 0, 1)^T$. The intensity distribution of this light across the $x_1x_2$-plane is described by the function $f \colon \sourceDomain\subset\mathbb{R}^2 \to \mathbb{R}$. Directly above the light source, there is a reflector. The geometry of the reflector's surface is defined by the function $u \colon \sourceDomain \to \mathbb{R}$, such that the height of the reflector is $x_3 = u(x_1, x_2)$. The reflector is considered to be a perfect mirror. That is, the angle of incidence---the angle at which light strikes the surface---equals the angle of reflection. Consequently, a light ray, emitted from the light source at a point $(x_1, x_2, 0)$, will be reflected at the point $(x_1, x_2, u(x_1, x_2))$ on the reflector. It will then travel in the direction $\mathbf{t} = \mathbf{s} - 2(\mathbf{s}\cdot\mathbf{n})\mathbf{n} = (t_1, t_2, t_3)^T$, where $\mathbf{n}$ is the unit normal of $u$ at $(x_1, x_2)$.

\begin{figure}
\centering
\includegraphics[width=0.7\linewidth]{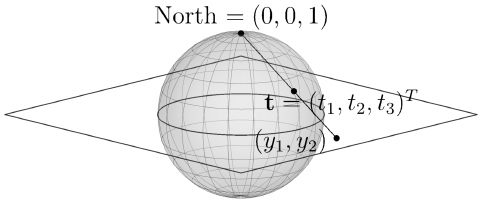}
% \begin{asy}
% import graph3;

% currentprojection = orthographic((5, 5, 1.8));

% triple north = (0, 0, 1);
% triple r = (0.0, 2.0, 1.0);
% r = r / length(r);
% triple proj = (r.x / (1 - r.z), r.y / (1 - r.z), 0.0);

% size(8cm,0);
% path3 myarc = rotate(18,Z) * Arc(c=O,
% normal=X, v1=-Z, v2=Z, n=10);
% surface backHemisphere =
% surface(myarc, angle1=0,
% angle2=180, c=O, axis=Z, n=10);
% surface frontHemisphere = surface(myarc, angle1=180,
% angle2=360, c=O, axis=Z, n=10);
% draw(backHemisphere, surfacepen=material(diffusepen=gray(1.0)+opacity(0.15), shininess=0.0), meshpen=gray(0.1)+opacity(0.3));
% draw(O--X, blue+linewidth(1pt)+opacity(0.0));
% draw(frontHemisphere, surfacepen=material(diffusepen=gray(1.0)+opacity(0.15), shininess=0.0), meshpen=gray(0.1)+opacity(0.5));
% draw(circle(c=(0, 0, 0), r=1, normal=Z), gray(0.2));

% dot(north);
% label("North $= (0, 0, 1)$", north + (0, 0, 0.199));

% dot(r);
% label("$\mathbf{t} = (t_1, t_2, t_3)^T$", r+(0.0, 0.8, 0.0));

% dot(proj);
% label("$(y_1, y_2)$", proj+(0.5, -0.28, 0.0));

% draw(north--proj);

% draw(plane((4.0, 0.0, 0.0), (0.0, 4.0, 0.0), O=(-2, -2, 0.0)), gray(0.2));
% \end{asy}
\caption{Stereographic projection of the reflected ray. }
\label{fig:stereo-coords}
\end{figure}

We assume a far-field target. That is, we assume the reflector's size to be negligible compared to the distance traveled by the reflected light. This assumption allows us to treat all reflected rays as having the same origin point. Therefore, the reflected rays are distinguished only by their direction vectors $\mathbf{t}$. These vectors can be visualized as points on a unit sphere. As such, each direction vector can be represented by just two coordinates, $y_1$ and $y_2$, using a stereographic projection from the north pole, as illustrated in Figure \ref{fig:stereo-coords}. Finally, at these projected coordinates $(y_1, y_2)$, the desired light distribution is determined by the function $g \colon \mathbb{R}^2 \to \mathbb{R}$.

The problem we wish to solve is as follows: given the functions $f$ and $g$, find the corresponding reflector height function $x_3 = \pdeSol(x_1, x_2)$ such that a light source emitting parallel light rays according to the distribution $f$ results in a reflected light distribution $g$. Based on this problem, we can derive

\begin{subequations}\label{eq:subeqs}
    \label{eq:transport-ma}
    \begin{equation}
        \label{eq:transport-ma1}
        \det\left(D^2 u(x_1, x_2)\right) = \frac{f(x_1, x_2)}{g(\nabla u(x_1, x_2))}, \quad (x_1, x_2) \in \mathcal{S},
    \end{equation}
    \noindent where $\mathcal{S}$ is the domain of $f$ and $u$, $D^2 u$ denotes the Hessian of $u$, and $\nabla u(x_1, x_2) = (u_{x_1}(x_1, x_2), u_{x_2}(x_1, x_2))^T$ is the gradient of $u$ at $(x_1, x_2)$. Following from the law of reflection, this gradient corresponds to the ray reflection in stereographic coordinates $(y_1, y_2)^T$ discussed above. As such, we also refer to the gradient of the reflector as the `mapping' of the reflector, as it determines where source light rays are sent to. This equation is an instance of the Monge-Amp\`ere equation. This equation is a consequence of local energy conservation. 
    
    Furthermore, all light emitted from the light source must be reflected by the reflector and arrive within the target domain $\mathcal{T}$ of $g$. As $f(x_1, x_2)$ must be greater than $0$ for all $(x_1, x_2)\in\sourceDomain$, $\det\left(D^2 u(x_1, x_2)\right)$ must be positive everywhere, which suggests $u$ should be either strictly convex (symmetric positive definite Hessian) or strictly concave (symmetric negative definite). We will assume convexity here. As such, the gradient of $u$ is a diffeomorphism, and we can simplify the condition to 
    \begin{equation}
        \label{eq:transport-boundary-condition}
    	\nabla u(\partial \mathcal{S}) = \partial \mathcal{T},
    \end{equation}
\end{subequations}

\noindent where $\partial \mathcal{S}$ and $\partial \mathcal{T}$ are the boundaries of $\mathcal{S}$ and $\mathcal{T}$, respectively. This condition is known as the transport boundary condition \cite{froese2012numerical}. 

Our goal is to find a numerical solution to this problem using artificial neural networks, similar to the approach presented by \citet{nystrom2023solving} for the Monge-Amp\`ere equation with Dirichlet boundary conditions, but extended to work for the transport boundary condition. We will first provide some additional relevant background relating to the methods we employed and the analyses we performed. 

\subsection{Artificial neural networks}
\label{sec:ann}
\begin{figure*}
\centering
\begin{tikzpicture}[x=2.2cm,y=1.4cm]
  \message{^^JNeural network with arrows}
  \readlist\Nnod{2,3,3,3,1} % array of number of nodes per layer
  
  \message{^^J  Layer}
  \foreachitem \N \in \Nnod{ % loop over layers
    \edef\lay{\Ncnt} % alias of index of current layer
    \message{\lay,}
    \pgfmathsetmacro\prev{int(\Ncnt-1)} % number of previous layer
    \foreach \i [evaluate={\y=\N/2-\i; \x=\lay; \n=\nstyle;}] in {1,...,\N}{ % loop over nodes
      
      % NODES
      % \ifthenelse{\N=0}{
      %   \node[node \n] (N\lay-\i) at (\x+6cm,\y) {$k_\i^{(\prev)}$};
      % }{
        \node[node \n] (N\lay-\i) at (\x,\y) {$a_\i^{(\prev)}$};
      % }
      %\node[circle,inner sep=2] (N\lay-\i') at (\x-0.15,\y) {}; % shifted node
      %\draw[node] (N\lay-\i) circle (\R);
      
      % CONNECTIONS
      \ifnum\lay>1 % connect to previous layer
        \foreach \j in {1,...,\Nnod[\prev]}{ % loop over nodes in previous layer
          \draw[connect arrow] (N\prev-\j) -- (N\lay-\i); % connect arrows directly
          %\draw[connect arrow] (N\prev-\j) -- (N\lay-\i'); % connect arrows to shifted node
        }
      \fi % else: nothing to connect first layer
      
    }
    
  }
  
  % LABELS
  \node[above=of N1-1,align=center,mygreen!60!black] at (N1-1.90) {Input\\[-0.2em]layer\\[-0.2em]$l=0$};
  \node[above=of N3-1,align=center,myblue!60!black] at (N3-1.90) {Hidden layers\\[-0.2em]$1\leq l\leq 3$};
  \node[above=of N5-1,align=center,myred!60!black] at (N\Nnodlen-1.90) {Output\\[-0.2em]layer\\[-0.2em]$l=L=4$};
  
\end{tikzpicture}
\caption{Example of a fully-connected feedforward neural network. Adapted from \citet{tikzNeuralNetworks}. }
\label{fig:fcn_diagram}
\end{figure*}
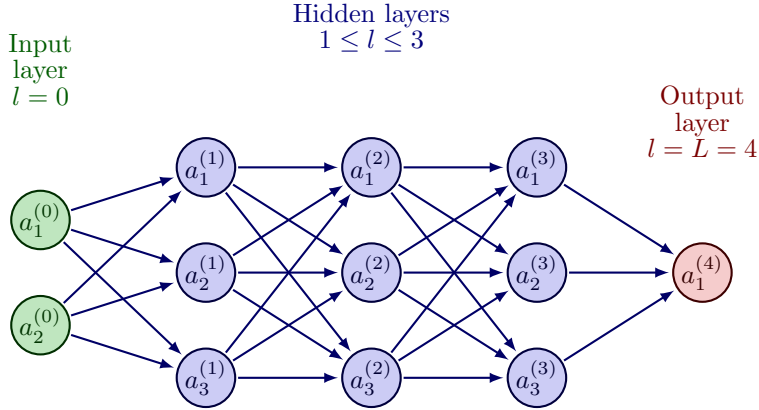
We can solve the PDE described in the previous section by representing its solution as a neural network and minimizing an appropriate loss function. Artificial neural networks (ANNs) are a class of machine learning models inspired by the structure of biological neural networks. Nowadays, there are a great variety of different ANN architectures, but as the function to be approximated in this particular problem is a scalar function of two variables, standard feedforward neural networks---also known as multilayer perceptrons (MLPs)---would seem most appropriate. An MLP consists of layers of neurons (units or nodes), where each neuron is connected to all neurons in the previous layer with weighted connections and biases. Each neuron computes its output by applying a nonlinear activation function to the sum of its inputs' weighted outputs and a bias.

The first layer of an MLP is referred to as the input layer, and the last layer is referred to as the output layer. All layers in between are known as hidden layers. The number of neurons in the input layer is equal to the number of inputs of the function to be approximated, and the number of neurons in the output layer is equal to the number of outputs of the function to be approximated. The number of hidden layers ($L-2$, excluding the input and output layers) and the number of neurons in each hidden layer ($n^{(l)}$, where $l \in \{1, \ldots, L-1\}$) are hyperparameters of the model. The weights between layers are stored in weight matrices $\mathbf{W}^{(l)}$, where the element $w_{ij}^{(l)}$ is the weight of the connection between the $i$-th neuron in layer $l-1$ and the $j$-th neuron in layer $l$. The biases of each layer are stored in bias vectors $\mathbf{b}^{(l)}$, where the element $b_i^{(l)}$ is the bias corresponding to the $i$-th neuron in layer $l$. Both weights and biases must be optimized during training.

In summary, the output of an MLP is computed as

\begin{align*}
    \mathbf{a}^{(0)} &= \mathbf{x}, \\
    \mathbf{z}^{(l)} &= \mathbf{W}^{(l)} \mathbf{a}^{(l-1)} + \mathbf{b}^{(l)}, & \quad 1 \leq l \leq L \\
    a^{(l)}_i &= \sigma(z^{(l)}_i), & \quad 1 \leq l \leq L \quad 1 \leq i \leq n^{(l)}, 
\end{align*}

\noindent where $\mathbf{x}$ is the input vector, $\mathbf{a}^{(l)}$ is the output vector of layer $l$, $\sigma(z)$ is some (non-linear) function referred to as the `activation' function, $n^{(l)}$ is the number of neurons in layer $l$, and $\mathbf{z}^{(l)}$ is the weighted sum of the outputs $\mathbf{a}^{(l-1)}$ of layer $l-1$ plus the bias $\mathbf{b}^{(l)}$ of layer $l$. The output of the MLP is then $\mathbf{a}^{(L)}$. 

According to the universal approximation theorem\cite{HORNIK1989359}, multilayer perceptrons with just a single hidden layer can approximate any Borel measurable function to any desired accuracy, provided a hidden layer of sufficient size and some mild assumptions on the activation function used. As such, a sufficiently large network should be capable of approximating the solution to the Monge-Amp\`ere equation. To do so, we must minimize an appropriate loss function, which we shall discuss in Section \ref{sec:transport-loss}.

\subsection{Least-squares solver}
\label{sec:background-lss}
While the approach we propose here represents the approximate solution of the Monge-Amp\`ere equation as a neural network, a more traditional discretized approach was presented by \citet{prins2015least}. Specifically, the gradient of $\pdeSol$ is approximated on a finite number of points, and the Hessian of $\pdeSol$---which corresponds to the Jacobian of these gradients---is determined via finite differences. In the context of optimal transport, the gradient of $\pdeSol$ represents a mapping, which we will refer to as $\mappingSol$ and its approximation as $\mappingApprox$. For the least-squares solver presented by \citet{prins2015least}, $\params\in\mathbb{R}^{n\times2}$ refers to the approximated mapping at each point on the grid, where $n$ denotes the number of points at which the mapping is approximated. 

The Jacobian of the mapping is defined as follows:

\begin{equation}
    D\mappingSol = 
    \begin{pmatrix}
        \frac{\partial m_1}{\partial x_1} & \frac{\partial x_2}{\partial y} \\
        \frac{\partial m_2}{\partial x_1} & \frac{\partial m_2}{\partial x_2}
    \end{pmatrix}
    = D^2\pdeSol. 
\end{equation}

\noindent This matrix should be equal to a real symmetric positive semidefinite matrix $\mathbf{P}(x_1, x_2)$ satisfying

\begin{equation}
    \det{\mathbf{P}(x_1, x_2)} = \frac{\sourceDist(x_1, x_2)}{\targetDist(\mappingSol(x_1, x_2))}
\end{equation}

\noindent for all $(x_1, x_2)\in\sourceDomain$. $\mathbf{P}(x_1, x_2)$ represents the Hessian of the reflector surface, which is why it should be symmetric, i.e. $\partial m_1 /\partial x_2 = \partial m_2 /\partial x_1$. In order to satisfy the above equality, the algorithm strives to minimize the functional

\begin{equation}
    J_I(\mappingSol, \mathbf{P}) = \tfrac{1}{2} \iint_{\sourceDomain}{\|D\mappingSol - \mathbf{P}\|^2 \, \diff x_1 \diff x_2}, 
\end{equation}

\noindent where $\|\cdot\|$ is the Frobenius norm. Here, $\mappingSol$ is minimized over the set $\mathcal{V} = \left[C^2(\sourceDomain)\right]^2$, i.e., the set of two-dimensional, twice continuously differentiable vector fields. Furthermore, $\mathbf{P}$ is minimized over

\begin{equation}
    \mathcal{P} = \left\{\mathbf{P}\in \left[C^1(\sourceDomain)\right]_{\text{spsd}}^{2\times 2} \mid \det\left(\mathbf{P}(x_1, x_2)\right) = \frac{f(x_1, x_2)}{g\left(\mappingSol(x_1, x_2)\right)}\right\},
\end{equation}

\noindent where $\left[C^1(\sourceDomain)\right]_{\text{spsd}}^{2\times 2}$ denotes all symmetric positive semidefinite $2\times 2$ matrices of $C^1(\sourceDomain)$-functions.

In addition to the functional $J_I$, the algorithm also has to satisfy the transport boundary condition defined in Equation (\ref{eq:transport-boundary-condition}). To this end, the algorithm aims to minimize the following functional as well:

\begin{equation}
    J_B(\mappingSol, \mathbf{b}) = \tfrac{1}{2} \oint_{\partial \sourceDomain} |\mappingSol - \mathbf{b}|^2 \, \diff s.
\end{equation}

The algorithm minimizes $J_B$ over the set:

\begin{equation}
    \mathcal{B} = \left\{ \mathbf{b} \in [C^1(\partial \sourceDomain)]^2 \mid \mathbf{b}(x) \in \partial \targetDomain \quad \forall x \in \partial \sourceDomain \right\}.
\end{equation}

These two functionals are then combined into

\begin{equation}
    J(\mappingSol, \mathbf{P}, \mathbf{b}) = (1 - \alpha)J_b(\mappingSol, \mathbf{b}) + \alpha J_I(\mappingSol, \mathbf{P}),
\end{equation}

\noindent where $\alpha$ is a parameter between $0$ and $1$ that controls the weight put on either functional.

The algorithm minimizes this functional by repeatedly performing three minimization steps for the boundary points $\mathbf{b}$, Jacobians $\mathbf{P}$, and mapping $\mappingSol$:

\begin{align}
    \mathbf{b}^{n+1} &= \underset{\mathbf{b} \in \mathcal{B}}{\mathrm{argmin}} \ J_B(\mappingSol^n, \mathbf{b}), \\
    \mathbf{P}^{n+1} &= \underset{\mathbf{P}\in\mathcal{P}}{\mathrm{argmin}} \ J_I(\mappingSol^n, \mathbf{P}), \\
    \mappingSol^{n+1} &= \underset{\mappingSol\in\mathcal{V}}{\mathrm{argmin}} \ J(\mappingSol, \mathbf{P}^{n+1}, \mathbf{b}^{n+1}).
\end{align}

\noindent For more information on how each of these minimization steps is performed, we refer the reader to \citet{prins2015least}.

The previously described steps are then repeated a number of times until some convergence criterion is met, yielding an approximation of the mapping $\mathbf{m}$. The reflector function would then be the function whose gradient matches the approximated mapping $\mathbf{m}$. However, as the previously described algorithm is not guaranteed to be the gradient of a function---that is, it need not be a conservative vector field---we instead try to find a function whose gradient is close to $\mathbf{m}$ in a least-squares sense, i.e.,

\begin{equation}
u = \underset{\psi\in C^2(\sourceDomain)}{\mathrm{argmin}} \, I(\psi), \quad \text{where }I(\psi) := \tfrac{1}{2} \iint_{\sourceDomain} |\nabla \psi - \mappingSol|^2 \, \diff x_1\, \diff x_2.
\end{equation}

This then yields the final approximation of the reflector $\pdeSol$. 

\section{Methods}
\label{sec:methods}
In order to find the weights and biases of an MLP that solves the Monge-Amp\`ere equation with the transport boundary condition, we must determine some loss function---to represent how well our MLP represents the solution of the PDE---and an optimization procedure---to minimize this loss. In the following section, we formulate the former. This is a modification of the loss function by \citet{nystrom2023solving}. Section \ref{sec:optimization} then describes how we went about minimizing this loss. 

\subsection{Modified loss function}
\label{sec:transport-loss}
To solve Equation (\ref{eq:transport-ma}), we need to find a convex function \( u(x_1, x_2) \) that satisfies the equation and the boundary conditions. We can approximate the solution to this equation using a neural network \( \pdeApprox(x_1, x_2) \) with parameters \( \params = \{\mathbf{W}^{(1)},\dots,\mathbf{W}^{(L)}, \mathbf{b}^{(1)},\dots,\mathbf{b}^{(L)}\} \). We can then define a loss function that penalizes the deviation of the approximated solution from the true solution. This loss function should satisfy the following requirements:

\begin{enumerate}
    \item The residual between the determinant of the Hessian of the approximated solution and the ratio \( \frac{\sourceDist(x_1, x_2)}{\targetDist(\nabla \pdeSol(x_1, x_2))} \) must be penalized in the interior of \( \sourceDomain \).
    \item The distance between the gradient of the approximated solution and target boundary \( \partial \targetDomain \) must be penalized on the boundary \( \partial \sourceDomain \).
    \item Non-convexity of the approximated solution must be penalized. That is, the eigenvalues of the Hessian of the approximated solution must be penalized if they are negative.
\end{enumerate}

We can define this loss function for any point \( (x_1, x_2) \in \sourceDomain \) and any set of parameters \( \params \) of some approximator \( \pdeApprox(x_1, x_2) \). Specifically, to satisfy the first requirement, we use the squared residual between the Hessian determinant of the approximated solution and the given function \( \frac{\sourceDist(x_1, x_2)}{\targetDist(\nabla u(x_1, x_2))} \):

\begin{equation}
    \label{eq:loss-interior-squared}
    L_I(x_1, x_2; \params) = \left( \det(D^2 \pdeApprox(x_1, x_2)) - \frac{f(x_1, x_2)}{g(\nabla \pdeApprox(x_1, x_2))} \right)^2. 
\end{equation}

For the second requirement, we wish all source boundary points to be mapped to the target boundary. To this end, we define the loss term.

\begin{equation*}
    \label{eq:boundary-loss-mlp}
    % L_{\text{B}}(x_1, x_2; \params) = \begin{cases}
    % \min_{(u, v)\in\partial\mathcal{T}}{\norm{\mappingApprox(x_1, x_2) - (u, v)}} & \text{if }(x_1, x_2) \text{ is on }\partial\sourceDomain, \\
    % 0 & \text{otherwise}
    % \end{cases}. 
    L_{B}(x_1, x_2; \params) = 
    \min_{(y_1, y_2)\in\partial\mathcal{T}}{\norm{\nabla\pdeApprox(x_1, x_2) - (y_1, y_2)}}^2,
\end{equation*}

\noindent for any point $(x_1, x_2)$ on $\partial\sourceDomain$. 

Depending on the shape of the target boundary $\partial\targetDomain$, the distance between a mapped point $\mappingApprox(x_1, x_2)$ and the target boundary can be difficult to compute analytically. In such a case, we could either compute the distance numerically or we could simply use a different function that is easier to compute. As long as this function attains its minimum on the target boundary, minimization should push the mapped source boundary points to the target boundary. 

The third requirement---that nonconvexity of the approximated solution must be penalized---is the same as that presented by \citet{nystrom2023solving}. As such, we use the same loss term as they did. To satisfy the third requirement, we must penalize the eigenvalues of the Hessian of the approximated solution if they are negative. In the two-dimensional case, the eigenvalues of the Hessian of the approximated solution are given by

\begin{equation*}
    \label{eq:eigenvalues}
    \lambda = \frac{h_{11}+h_{22}\pm\sqrt{(h_{11}+h_{22})^2-\det\left(D^2u_{\boldsymbol{\theta}}\right)}}{2},
\end{equation*}

\noindent where \( h_{ij} \) are the entries of the Hessian matrix \( D^2u_\theta \). As we are assuming \( f \) to be nonnegative, and as Equation (\ref{eq:loss-interior-squared}) already penalizes the residual between the determinant of the Hessian of the approximated solution and the given function \( f \), we know the eigenvalues will be nonnegative when \( h_{11} + h_{22} \geq 0 \). Thus, we only need to penalize the negativity of this sum, leading to the loss term

\begin{equation}
    \label{eq:loss-convexity}
    L_C(x_1, x_2; \params) = \min \left(h_{11} + h_{22}, 0\right)^2.
\end{equation}

\begin{figure}
    \centering
    \includegraphics[width=0.6\textwidth]{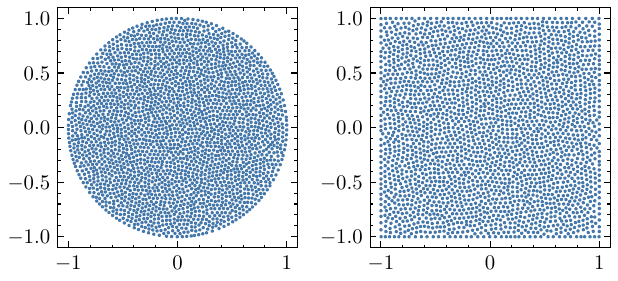}
    \caption{The $2500$ interior points used for optimization for the unit circle and square. The unit circle is used for Problems \protect\problemreff{A}, \protect\problemreff{B}, \protect\problemreff{C}, and \protect\problemreff{E}. The unit square is used only for \protect\problemref{D}. }
    \label{fig:opt-points}
\end{figure}

As we wish to minimize \( L_B \), \( L_I \), and \( L_C \) for all points \( \textbf{x} \in \sourceDomain \), we can define the loss function \( \mathcal{L} \) as

\begin{equation}
    \label{eq:total-loss}
    \mathcal{L}(\params) = \int_\sourceDomain \left(\alpha L_I(\textbf{x}, \params) + \beta L_C(\mathbf{x}, \params) \right)\,\diff\mathbf{x} + \int_{\partial \sourceDomain} \gamma L_B(\mathbf{b}, \params)\,\diff\mathbf{b},
\end{equation}

\noindent where \( \alpha \), \( \beta \), and \( \gamma \) are constants that control the relative importance of each term. As we cannot directly compute the integral in Equation (\ref{eq:total-loss}), we must instead approximate it. We can accomplish this by sampling a set of interior points \( \textbf{x}_1, \ldots, \textbf{x}_N \) from \( \mathcal{S} \) and boundary points \( \textbf{b}_1, \ldots, \textbf{b}_M \) from \( \partial\mathcal{S} \), and redefining the loss as

\begin{equation}
    \label{eq:total-loss-approx}
    \mathcal{L}(\boldsymbol{\theta}) \appropto \frac{1}{N} \sum_{i=1}^N \left(\alpha L_I(\textbf{x}_i, \params)\right) + \beta L_C(\textbf{x}_i, \params) + \frac{1}{M} \sum_{i=1}^M \gamma L_B(\textbf{b}_i, \params),
\end{equation}

\noindent where \( N \) and \( M \) are the number of interior and boundary points sampled. We can then minimize this loss function using gradient descent, or any other optimization algorithm. Each point $\mathbf{x_i}$ could be sampled randomly and independently from the source domain $\sourceDomain$, as performed by \citet{nystrom2023solving}, or we could use a grid, such as is necessary for the least-squares solver. However, we instead used Poisson disk sampling \cite{bridson2007fast} to obtain points more evenly distributed throughout the source domain, see Figure \ref{fig:opt-points}. 

The optimal selection of hyperparameters $\alpha$, $\beta$, and $\gamma$ remains an open question. Although adaptive methods for tuning these values during optimization have been proposed to enhance the performance of MLP-based approaches to solve PDEs (e.g. \cite{xiang2022self}), our empirical results suggest that a simpler approach is often sufficient. We have found that setting $\alpha = \beta = \gamma = 1$ consistently yields satisfactory performance across the range of problems examined in this study. Consequently, we have employed these values for all results presented here. 

\subsection{Optimization}
\label{sec:optimization}
In order to find the necessary weights and biases for a neural network that minimizes the loss function described above, we must utilize some optimizer. Optimizers commonly used for modern machine learning applications---such as Adam \cite{kingma2014adam}---can be robust in the presence of the stochasticity normally present in these same applications. However, while large-scale machine learning applications generally require stochasticity, it is not strictly necessary for the problem considered here. Furthermore, we find such methods struggle to converge within reasonable time frames and to a reasonable level of accuracy when optimizing the loss function described in the previous section, compared to what one might expect from more traditional PDE solvers. Optimizers like Adam rely solely on gradient evaluations of the loss function and do not consider the Hessian. While the computation of the full Hessian can be quite computationally and memory-intensive---rendering it infeasible for larger neural networks---quasi-Newton methods such as the Broyden–Fletcher–Goldfarb–Shanno (BFGS)\cite{fletcher2000practical} algorithm and more specifically the limited-memory BFGS\cite{liu1989limited} allow us to approximate the Hessian while incurring minimal performance and memory overhead. As such, we implemented this optimization algorithm using the linesearch algorithm proposed by \citet{more1994line} and used it for all experiments presented here. Furthermore, we contrast its performance with the aforementioned Adam optimizer as well to demonstrate the effectiveness of quasi-Newton methods over first-order optimizers for this problem. 

\section{Numerical examples}
\label{sec:results}
We will now present five numerical examples to demonstrate the performance of the proposed method. For each of these examples, we will compare the performance of our method against that of the least-squares solver described in Section \ref{sec:background-lss}. All experiments performed here were executed on a desktop with an AMD Ryzen 5 3600 CPU, an NVIDIA GTX 1060 6 GB GPU, and 24 GB of RAM. An existing implementation of the least-squares solver written in MATLAB was used, and the experiments presented here were performed using MATLAB R2023b specifically. We implemented the neural network-based method in Python using JAX \cite{jax2018github} for the requisite tensor computations, automatic differentiation, and GPU acceleration. In the following sections, we will refer to the least-squares solver as LSS and the MLP-based method as MLP-L-BFGS.

\begin{table}[h]
\centering
\begin{subtable}[t]{1.0\textwidth}
\centering
\begin{tabular}{@{}c c c@{}}
\toprule
 & \multicolumn{2}{c}{\textbf{Solver}} \\
\cmidrule(lr){2-3}
& \textbf{LSS} & \textbf{MLP-L-BFGS} \\
\midrule
Problem A & \(4.590 \times 10^{-4}\) & \(2.823 \times 10^{-6}\) \\
Problem B & \(1.846 \times 10^{-3}\) & \(2.403 \times 10^{-6}\) \\
Problem C & \(1.331 \times 10^{-4}\) & \(1.204 \times 10^{-6}\) \\
\bottomrule
\end{tabular}
\caption{Normalized mean absolute error between the exact and computed solutions for Problems A, B, and C. }
\label{subtable:mean-l2-error}
\end{subtable}

\vspace{1cm}

\begin{subtable}[t]{1.0\textwidth}
\centering
\begin{tabular}{@{}c c c@{}}
\toprule
& \multicolumn{2}{c}{\textbf{Solver}} \\
\cmidrule(lr){2-3}
& \textbf{LSS} & \textbf{MLP-L-BFGS} \\
\midrule
Problem D & \(2.145 \times 10^{-3}\) & \(1.508 \times 10^{-3}\) \\
Problem E & \(7.363 \times 10^{-2}\) & \(1.923 \times 10^{-2}\) \\
\bottomrule
\end{tabular}
\caption{Normalized mean absolute error between the exact and ray-traced results for Problems D and E. }
\label{subtable:relative-mean-absolute-error}
\end{subtable}

\caption{Obtained errors for the LSS and MLP-L-BFGS solvers for each problem. The first three problems are evaluated in terms of mean Normalized Mean Absolute Error compared against the known exact reflector surface, and the last two problems are evaluated in terms of the error between a ray-traced image and the exact target image.}
\label{table:problem-errors}
\end{table}

\needspace{8\baselineskip} % Adjust the space as needed
\subsection{Problem A: Symmetrical circle-to-circle}
\label{sec:prob1}
\defineproblem{A}{}
\begin{figure}[H]
    \centering
    \includegraphics[width=0.85\textwidth]{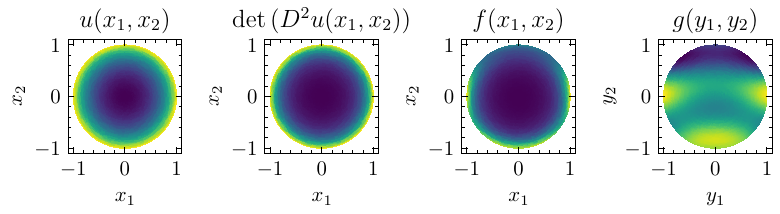}
    \caption{[\protect\problemref{A}] In order, $u(x_1, x_2)$, $\det(D^2 u(x_1, x_2))$, $f(x_1, x_2)$, and $g(y_1, y_2)$. }
    \label{fig:prob1-viz}
\end{figure}
\begin{figure*}
    \centering
    \begin{subfigure}{0.96\textwidth}
        \centering
        \includegraphics[width=0.66\textwidth]{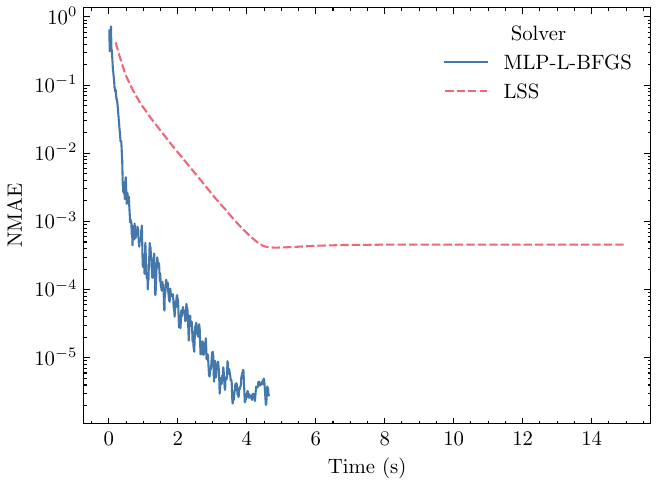}
        \subcaption{Normalized mean absolute error as a function of time. }
        \label{fig:prob1-convergence}
    \end{subfigure}
    \begin{subfigure}{0.96\textwidth}
        \centering
        \includegraphics[width=0.66\textwidth]{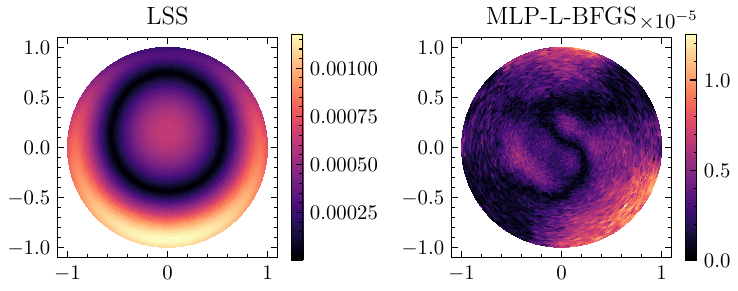}
        \subcaption{NMAE across the source domain $\sourceDomain$. }
        \label{fig:prob1-domain}
    \end{subfigure}
    \caption{[\protect\problemref{A}] Convergence and final error across the source domain $\sourceDomain$ for the Least-Squares Solver (LSS) and the proposed MLP-L-BFGS method in terms of NMAE. }
\end{figure*}
We first consider a simple rotationally symmetric reflector where $\partial\sourceDomain$ and $\partial\targetDomain$ are both circles. We chose the reflector $\pdeSol(x_1, x_2) = \frac{\exp\left(x_1^2 + x_2^2\right)}{2e}$ and the target light distribution $\targetDist(y_1, y_2) = \sin(y_1^2 + y_2) \cos(4y_2) + 2$. Using Equation (\ref{eq:transport-ma1}) we can then compute $\sourceDist(x_1, x_2)$, see Figure \ref{fig:prob1-viz}. We used a $100\times 100$ polar grid for optimization and evaluation of the. The LSS solver was terminated after $15$ seconds. 

For the MLP-based method, we used $2500$ interior points distributed evenly across the source domain using Poisson disk sampling, and $500$ boundary points distributed evenly on the source boundary. We used a network with three hidden layers, each with $32$ hidden neurons. For each hidden layer, we used the activation function $\sigma(z) = \tanh^2\left(z\right)$, which we have empirically found to generalize well across different problems. For the output layer, we used identity activation, that is, $\mathbf{a}^{(L)} = \sigma(\mathbf{z}^{(L)}) = \mathbf{z}^{(L)}$. Optimization of the MLP-L-BFGS method was terminated either when the linesearch subroutine failed to find an adequate stepsize within $16$ iterations or if the timeout of $15$ seconds was reached. Finally, we evaluated the resulting trained neural network on the same $100\times 100$ grid as the LSS. The metric we used to evaluate the method is the normalized mean absolute error (NMAE) between the exact reflector and the approximated reflector surface obtained by either LSS or MLP-L-BFGS solver. The NMAE is defined as

\begin{equation}
    \text{NMAE} = \frac{\frac{1}{N}\sum_{i=1}^{N} |\pdeSol(x_i, y_i) - \pdeApprox(x_i, y_i)|}{\frac{1}{N}\sum_{i=1}^{N} |\pdeSol(x_i, y_i)|},
\end{equation}

\noindent where $N$ is the total number of points in the evaluation grid. 

This error is shown as a function of time in Figure \ref{fig:prob1-convergence} for both solvers. The error across the source domain is shown in Figure \ref{fig:prob1-domain}. Furthermore, the final errors obtained for both methods are shown in Table \ref{table:problem-errors}. These results show that the MLP-L-BFGS solver achieves greater accuracy and converges more quickly than the LSS for this simple example. 

\subsection{Problem B: Asymmetrical circle-to-circle 1}
\label{sec:prob2}
\defineproblem{B}{}
\begin{figure}[H]
    \centering
    \includegraphics[width=0.85\textwidth]{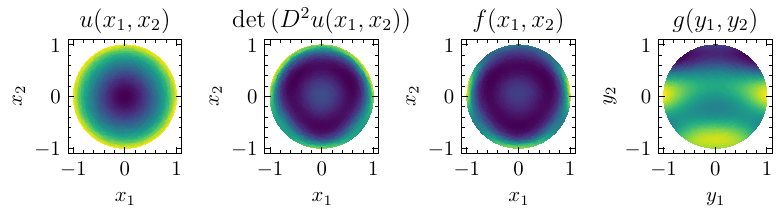}
    \caption{[\protect\problemref{B}] In order, $u(x_1, x_2)$, $\det(D^2 u(x_1, x_2))$, $f(x_1, x_2)$, and $g(y_1, y_2)$. }
    \label{fig:prob2-viz}
\end{figure}
\begin{figure*}
    \centering
    \begin{subfigure}{0.96\textwidth}
        \centering
        \includegraphics[width=0.66\textwidth]{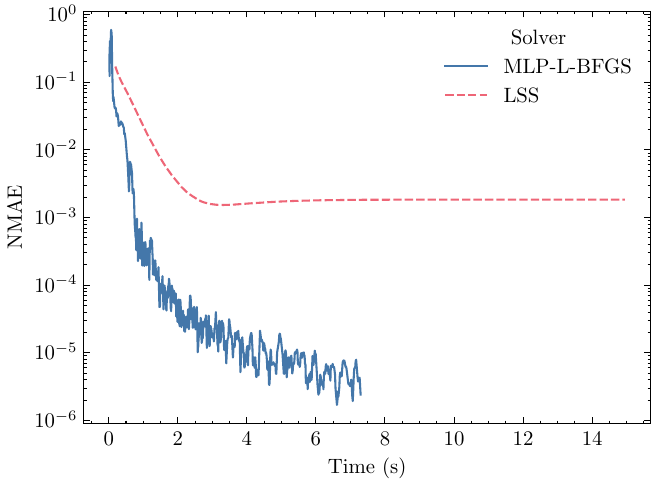}
        \subcaption{NMAE as a function of time for the Least-Squares Solver (LSS) and the proposed MLP-L-BFGS method. }
    \end{subfigure}
    \begin{subfigure}{0.96\textwidth}
        \centering
        \includegraphics[width=0.66\textwidth]{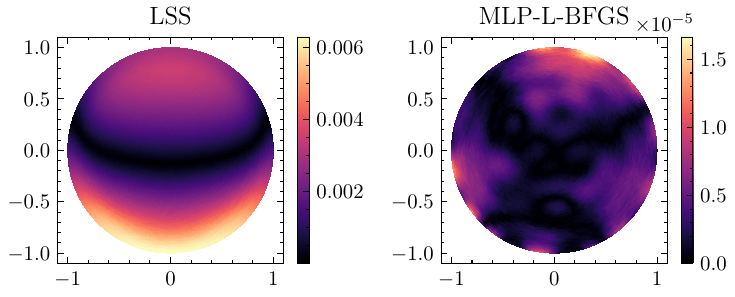}
        \subcaption{NMAE across the source domain $\sourceDomain$ for the Least-Squares Solver (LSS) and the proposed MLP-L-BFGS method. }
    \end{subfigure}
    \caption{[\protect\problemref{B}] Convergence and final error across the source domain $\sourceDomain$ for the Least-Squares Solver (LSS) and the proposed MLP-L-BFGS method in terms of NMAE. }
    \label{fig:problem2-results}
\end{figure*}
While a rotationally symmetric problem reduces to a 2D reflector problem---meaning we could employ simpler methods to solve it---our method can also be used to solve non-symmetric reflectors. For this problem, we still set $\partial\sourceDomain$ and $\partial\targetDomain$ to both be unit circles. However, we used the non-rotationally symmetric reflector

\begin{equation*}
\pdeSol(x_1, x_2) = \frac{1}{2} \left(x_1^2 + x_2^2\right) + \frac{\left(\frac{x_1^{2} x_2}{2} + \frac{\cos{\left(x_1 x_2 \right)}}{2}\right) \left(\cos{\left(\pi \left(x_1^{2} + x_2^{2}\right) \right)} + 1\right)}{2\pi^{2}}.
\end{equation*}

We chose the same target light distribution as the first problem, and once again used Equation (\ref{eq:transport-ma1}) to compute the source light distribution, see the visualization in Figure \ref{fig:prob2-viz}. We used the same optimization and evaluation parameters as described in the previous section. The results for both solvers are visualized in Figure \ref{fig:problem2-results}. The final errors obtained for both methods are shown in Table \ref{table:problem-errors}. Once again, the MLP-L-BFGS solver achieves greater accuracy and converges more quickly than the LSS. 

\subsection{Problem C: Asymmetrical circle-to-circle 2}
\label{sec:prob3}
\defineproblem{C}{}
\begin{figure}[H]
    \centering
    \includegraphics[width=0.85\textwidth]{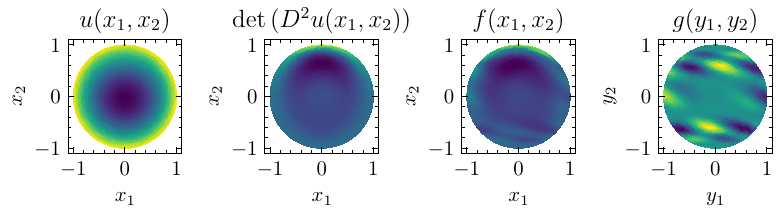}
    \caption{[\protect\problemref{C}] In order, $u(x_1, x_2)$, $\det(D^2 u(x_1, x_2))$, $f(x_1, x_2)$, and $g(y_1, y_2)$. }
    \label{fig:prob3-viz}
\end{figure}
\begin{figure*}
    \centering
    \begin{subfigure}{0.96\textwidth}
        \centering
        \includegraphics[width=0.66\textwidth]{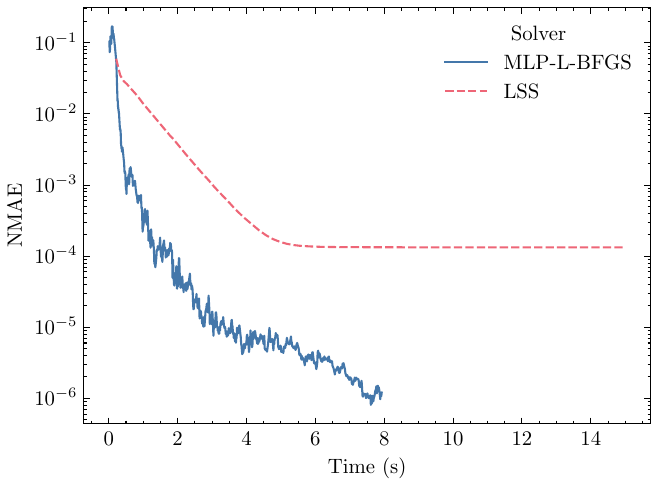}
        \subcaption{NMAE as a function of time for the Least-Squares Solver (LSS) and the proposed MLP-L-BFGS method. }
    \end{subfigure}
    \begin{subfigure}{0.96\textwidth}
        \centering
        \includegraphics[width=0.66\textwidth]{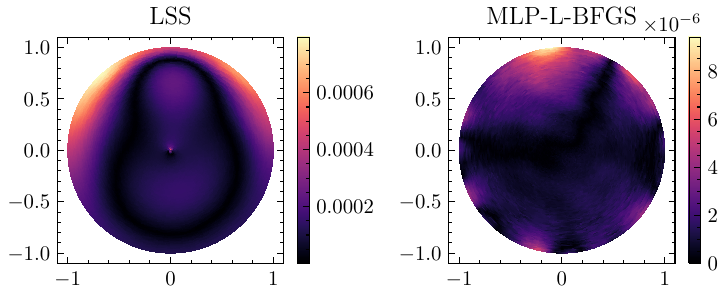}
        \subcaption{NMAE across the source domain $\sourceDomain$ for the Least-Squares Solver (LSS) and the proposed MLP-L-BFGS method. }
    \end{subfigure}
    \caption{[\protect\problemref{C}] Convergence and final error across the source domain $\sourceDomain$ for the Least-Squares Solver (LSS) and the proposed MLP-L-BFGS method in terms of NMAE. }
    \label{fig:problem3-results}
\end{figure*}
We evaluated another asymmetrical circle-to-circle problem with a slightly more complicated reflector and target distribution. Specifically, we used the reflector function

\begin{equation*}
    u(x_1, x_2) = \frac{1}{2} \left(x_1^2 + x_2^2\right) +\frac{\exp\left(-x_1^2 + x_2\right) \left(1 + \cos\left(\pi \left(x_1^2 + x_2^2\right)\right)\right)}{5\pi^2}, 
\end{equation*}

\noindent and the target distribution function

\begin{equation*}
    g(y_1, y_2) = \sin\left(y_1^2 + 8y_2^3\right) \cos\left(5y_2\right) \sin\left(5y_1 + 7y_2\right) + 3.
\end{equation*}

\noindent This is visualized in Figure \ref{fig:prob3-viz}. 

Optimization and evaluation were performed in the same way as for the previous two problems. The results of this are shown in Figure \ref{fig:problem3-results}. The final errors obtained for both methods are shown in Table \ref{table:problem-errors}. Although the target distribution here is somewhat more complex than those of \problemref{A} and \problemref{B}, we still see similar results for both solvers. However, there does appear to be a slight artifact at the center of the domain in the LSS error distribution, which may be related to the usage of polar coordinates for the LSS, which the MLP-L-BFGS method does not use. 

\subsection{Problem D: Square-to-circle}
\label{sec:prob4}
\defineproblem{D}{}
\begin{figure}[H]
    \centering
    \includegraphics[width=0.42\textwidth]{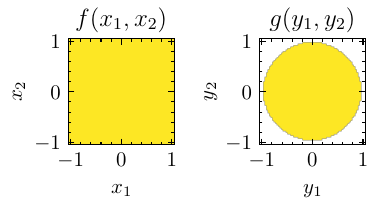}
    \caption{[\protect\problemref{D}] In order, $f(x_1, x_2)$ and $g(y_1, y_2)$. }
    \label{fig:prob4-viz}
\end{figure}
\begin{figure*}
    \centering
    \begin{subfigure}{0.96\textwidth}
        \centering
        \includegraphics[width=0.66\textwidth]{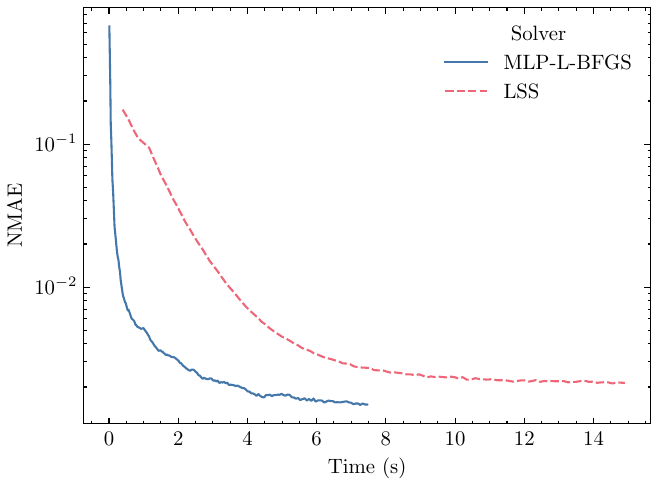}
        \subcaption{NMAE between the desired target distribution $\targetDist$ and the ray-traced result as a function of time for the Least-Squares Solver (LSS) and the proposed MLP-L-BFGS method. }
    \end{subfigure}
    \begin{subfigure}{0.96\textwidth}
        \centering
        \includegraphics[width=0.5\textwidth]{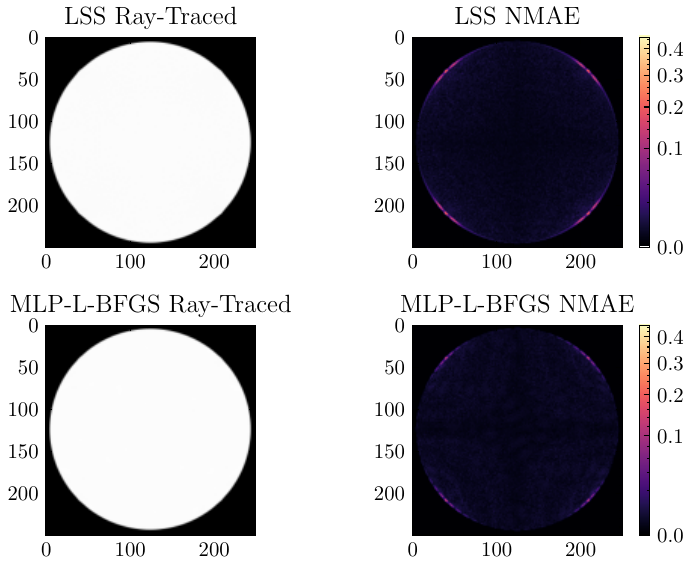}
        \subcaption{NMAE between the desired target distribution $\targetDist$ and the ray-traced result across the source domain $\sourceDomain$ for the Least-Squares Solver (LSS) and the proposed MLP-L-BFGS method. }
        \label{fig:prob4-domain}
    \end{subfigure}
    \caption{[\protect\problemref{D}] Convergence and final error across the source domain $\sourceDomain$ for the Least-Squares Solver (LSS) and the proposed MLP-L-BFGS method in terms of the normalized mean absolute error. }
    \label{fig:problem4-results}
\end{figure*}
While the previous three problems all mapped circular source domains to circular target domains, we can also use different shapes for both domains, as long as the source domain remains convex \cite{mohammed2008existence}. To demonstrate this, we present two such problems as well. Firstly, we consider the problem of mapping a square light source to a circular target. To keep things simple, we will use uniform light distributions across both domains. See Figure \ref{fig:prob4-viz}. 

As we do not have an analytic expression for such a reflector, we instead evaluate the performance of each solver using ray-tracing. That is, we sample quasi-random points from the source domain, and compute the reflection of each such ray onto the target, assigning it to a discrete bin. By aggregating a large number of rays, we obtain the approximate target distribution corresponding to the given source light distribution and the approximate reflector. We can then compare this image against the target image. For the MLP-L-BFGS method, we can simply compute the reflection by means of automatic differentiation. For the LSS, we instead interpolate the obtained mapping using cubic spline interpolation to obtain a mapping for arbitrary points within the source domain.

For optimization, we used the same parameters as specified previously. To evaluate each method using the ray-tracing approach, we used $250\times 250$ bins and $100$ million quasi-randomly sampled rays. We then computed the normalized mean absolute error between the obtained ray-traced image and the known target image, based on the known target distribution $\targetDist$. These results are shown in Figure \ref{fig:problem4-results}. The final numerical errors obtained for both methods are shown in Table \ref{table:problem-errors}. As we can see in Figure \ref{fig:prob4-domain}, both methods struggle to remove the corners of the source square for the circular output and they obtain a similar final accuracy, though the MLP-L-BFGS solver converges somewhat faster.

\subsection{Problem E: Circle-to-flower}
\label{sec:prob5}
\defineproblem{E}{}
\begin{figure}[H]
    \centering
    \includegraphics[width=0.42\textwidth]{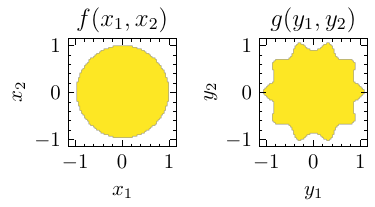}
    \caption{$f(x_1, x_2)$ and $g(y_1, y_2)$. }
    \label{fig:prob5-viz}
\end{figure}
\begin{figure*}
    \centering
    \begin{subfigure}{0.96\textwidth}
        \centering
        \includegraphics[width=0.66\textwidth]{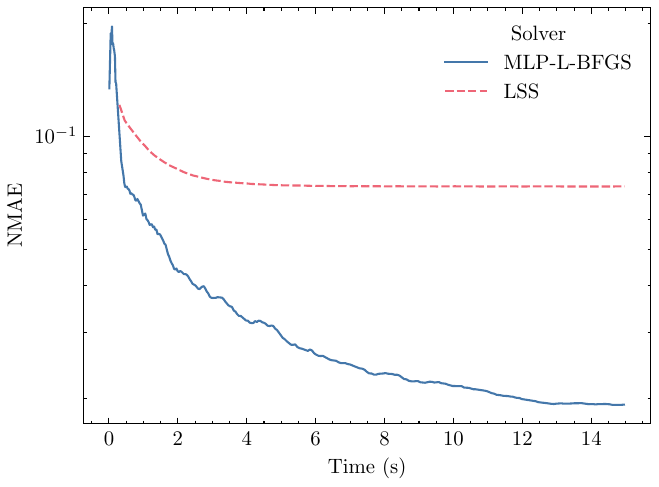}
        \subcaption{[\protect\problemref{E}] NMAE between the desired target distribution $\targetDist$ and the ray-traced result as a function of time for the Least-Squares Solver (LSS) and the proposed MLP-L-BFGS method. }
    \end{subfigure}
    \begin{subfigure}{0.96\textwidth}
        \centering
        \includegraphics[width=0.5\textwidth]{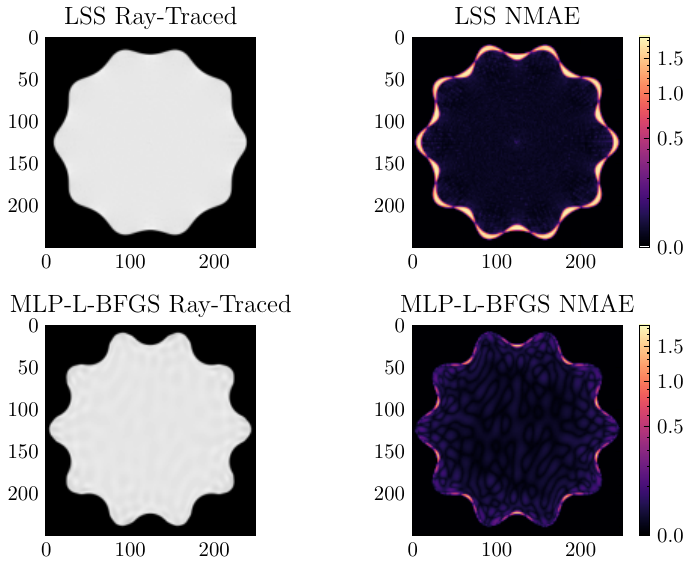}
        \subcaption{NMAE between the desired target distribution $\targetDist$ and the ray-traced result across the source domain $\sourceDomain$ for the Least-Squares Solver (LSS) and the proposed MLP-L-BFGS method. }
        \label{fig:problem5-results-2}
    \end{subfigure}
    \caption{[\protect\problemref{E}] Convergence and final error across the source domain $\sourceDomain$ for the Least-Squares Solver (LSS) and the proposed MLP-L-BFGS method in terms of the normalized mean absolute error. }
    \label{fig:problem5-results}
\end{figure*}
Finally, we evaluated mapping a uniform disk-shaped distribution to a uniform `flower'-shaped distribution. Specifically, we defined the target domain as

\begin{equation*}
    \targetDomain = \left\{(x_1, x_2) \,\middle|\, x_1^2 + x_2^2 \leq 1 + \frac{1}{10} \cos\left(10 \cdot \atantwo\left(x_2, x_1\right)\right)\right\}.
\end{equation*}

\noindent See the visualization in Figure \ref{fig:prob5-viz}. 

As we do not know the analytic expression for this reflector either, we once again used the ray-tracing approach to evaluate the accuracy of both solvers. These results are shown in Figure \ref{fig:problem5-results}. The final numerical errors obtained for both methods are shown in Table \ref{table:problem-errors}. Additionally, as we do not have an analytic expression for the shortest distance to the target boundary either, we instead used the loss term

\begin{equation*}
    L_{B}(x_1, x_2; \params) =
    \norm{(x_1, x_2) - \left(1+\frac{\cos(10\alpha)}{10}\right) \left(\cos\left(\alpha\right), \sin\left(\alpha\right)\right)}^2,
\end{equation*}

\noindent where $\alpha = \atantwo\left(x_2, x_1\right) = \arg\left(x_1 + i x_2\right)$ and $\norm{\cdot}$ is the Euclidean norm. We also experimented with using numerical minimization to find a closer approximation to the shortest distance. However, this did not yield significantly better accuracy and is more computationally expensive. As such, we used this simplified expression instead. 

We can see that both methods struggle even more with this problem than \problemref{D}. However, the final accuracy obtained by the MLP-L-BFGS solver is somewhat better, as well as converging slightly faster. Furthermore, if we look at the error distributions in Figure \ref{fig:problem5-results-2}, we can see that both methods' error is concentrated near the boundary, and the LSS in particular struggles to correctly represent the target boundary. 

\section{Hyperparameter study}
\label{sec:results-hyper}
While we kept all hyperparameters---the number of layers, number of hidden neurons, etc.---fixed for the results presented in Section \ref{sec:results}, the optimal choice of these parameters is likely to depend on the exact problem one is trying to solve. The choices we made for the results presented in Section \ref{sec:results} were chosen manually and somewhat arbitrarily. Through this, we hope to show that---as long as one picks somewhat `reasonable' hyperparameter settings---we can obtain sufficiently accurate results in a sufficiently small amount of time. To demonstrate this point more clearly, we will examine the impact of four different hyperparameters on Problems \problemreff{A}, \problemreff{B}, and \problemreff{C} in terms of NMAE. Specifically, we will consider the effects of the number of interior points, the number of boundary points, the number of hidden layers, and the number of hidden neurons per hidden layer. Additionally, we will show the effect of each hyperparameter on the convergence curve for \problemref{A}, to indicate the trade-off between computation time and accuracy for each hyperparameter. 

\begin{figure}
    \centering
    \includegraphics[width=0.97\textwidth]{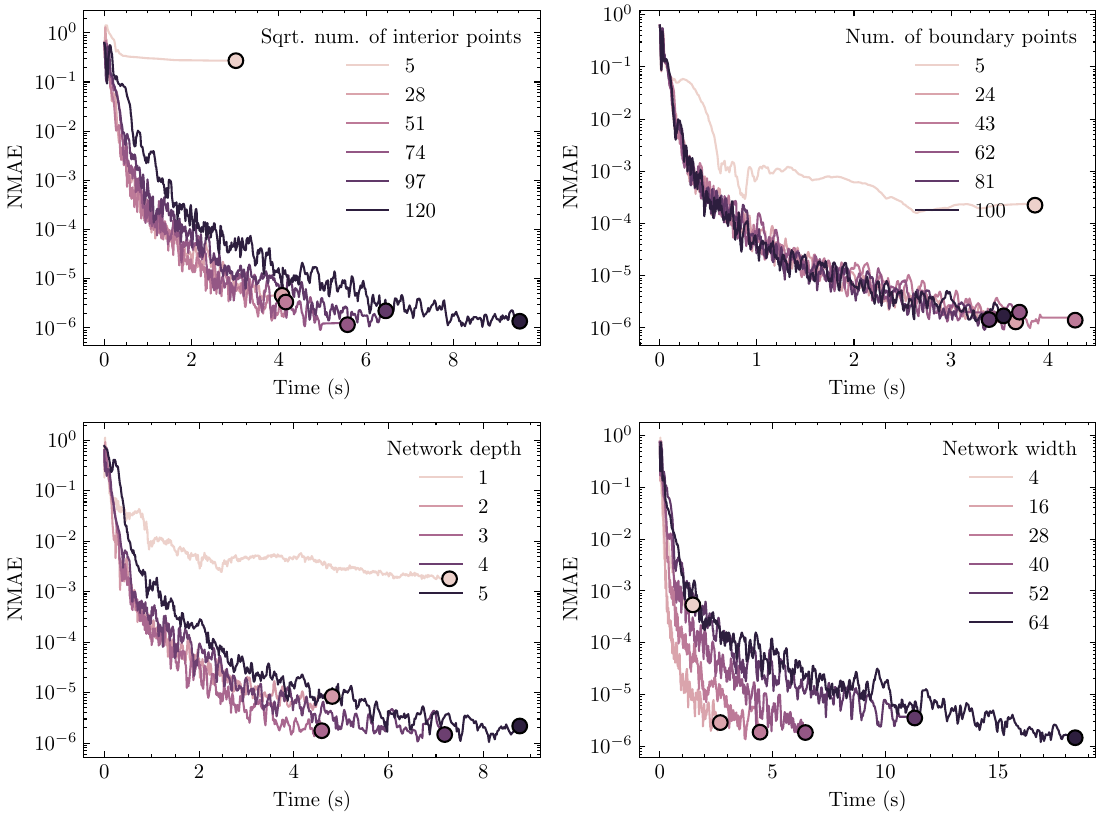}
    \caption{NMAE between the approximation $\pdeApprox$ and the exact $\pdeSol$ as a function of time when varying the number of interior points, the number of boundary points, network depth, and network width for Problem \protect\problemreff{A}. For each plot, the remaining hyperparameters are left unchanged from those specified in Section \ref{sec:results}. Each marker indicates the time and NMAE where the optimization terminated. }
    \label{fig:hyperparams-conv-prob1}
\end{figure}

\begin{figure}
    \centering
    \includegraphics[width=0.97\textwidth]{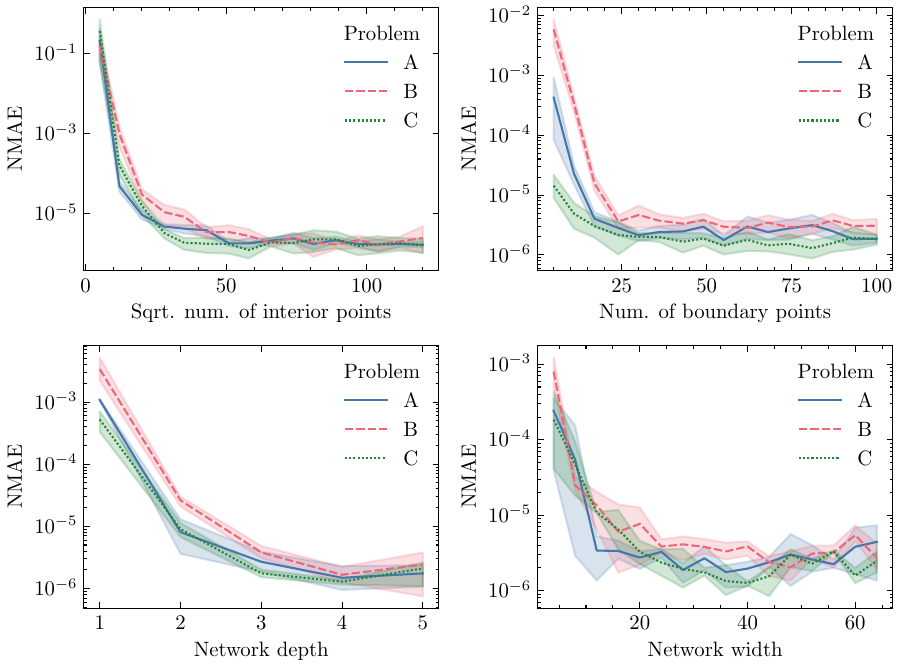}
    \caption{NMAE between the approximation $\pdeApprox$ and the exact $\pdeSol$ as a function of the number of interior, the number of boundary points, network depth, and network width for Problem \protect\problemreff{A}, \protect\problemreff{B}, and \protect\problemreff{C}. }
    \label{fig:hyperparams-n-interior}
\end{figure}

Firstly, we will examine the effect of the number of interior points. Together with the number of boundary points, the number of interior points controls the accuracy of the approximation of the first integral in Equation (\ref{eq:total-loss}). Increasing the number of interior points yields a more accurate approximation but is also more computationally expensive. We optimized the same neural network---with the same initialization scheme and consisting of three hidden layers of 32 neurons each---using a varying number of interior points. As network initialization is random, we used $10$ different seeds for each evaluation, which are visualized using a $95\%$ confidence interval. The resulting error as a function of this parameter is shown in the top left plot of Figure \ref{fig:hyperparams-n-interior}. Note that we plot the error as a function of the square-root of the number of interior points, as traditional methods would generally use a 2D grid. As we can see in this plot, using more points indeed yields greater accuracy, as we would expect. Additionally, it seems that we need considerably fewer points to obtain an accuracy similar to that of the LSS, and that this trend is the same for all three problems considered here. Furthermore, we can see in Figure \ref{fig:hyperparams-conv-prob1} that changing the number of interior points quickly leads to a more accurate result while slightly increasing the computation time, as we would expect. 

We also varied the number of boundary points. The results of this experiment are shown in the top-right plot of Figure \ref{fig:hyperparams-n-interior}. As we can see here, the effect of the number of boundary points is minimal, we only need a very small number of points to adequately enforce the transport boundary condition, and adding more points yields no further improvement in accuracy. However, it should be noted that all three problems shown here use circular source and target domains. We expect that more complex boundaries as well as a greater difference between the source and target domain boundaries would necessitate a greater number of boundary points. If we look at Figure \ref{fig:hyperparams-conv-prob1} again, we can see that increasing the number of boundary points quickly saturates the accuracy. In addition, we note that the computation time is not greatly affected by this hyperparameter. This can firstly be explained by the fact that this parameter scales linearly, whereas the number of interior points grows quadratically. Furthermore, the boundary loss function in Equation (\ref{eq:boundary-loss-mlp}) requires only the computation of the mapping at each point, whereas each interior point additionally requires the computation of the determinant of the Hessian. 

In addition to the number of interior and boundary points, we can also consider the effect of the size of the network. Specifically, we can consider the effect of the width---the number of neurons per hidden layer---and the depth---the number of hidden layers---on the performance of the network. Larger networks can represent more complex functions, but also result in greater computational overhead and may be more difficult to optimize. 

% \begin{figure}
%     \centering
%     \includegraphics[width=0.6\textwidth]{figures/results/hyperparams_network.pdf}
%     \caption{Normalized mean absolute error between the approximation $\pdeApprox$ and the exact $\pdeSol$ as a function of the number of hidden layers (depth) and number of hidden neurons per layer (width) for Problem A, B, and C. }
%     \label{fig:hyperparams-n-layers}
% \end{figure}

The bottom-left plot in Figure \ref{fig:hyperparams-n-interior} shows the effect of network depth on the NMAE. As we can see in this figure, a shallow network---only one hidden layer---fails to obtain an accurate solution. However, adding even a slight amount of depth already yields more acceptable accuracy. Figure \ref{fig:hyperparams-conv-prob1} once again shows how the accuracy quickly saturates, and computation increases only mildly with increasing depth. 

\begin{figure}
    \centering
    \includegraphics[width=0.55\textwidth]{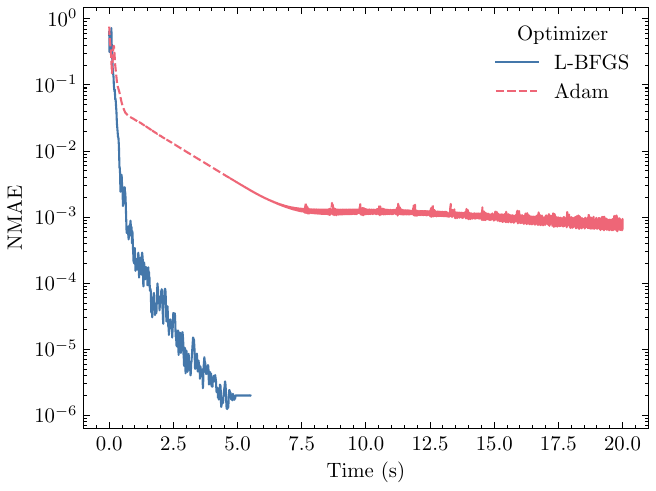}
    \caption{Normalized mean absolute error between the approximation $\pdeApprox$ and the exact $\pdeSol$ as a function of time for \protect\problemref{A} for Adam and L-BFGS optimization. }
    \label{fig:hyperparams-optimizer}
\end{figure}

The effect of the width of the network on final accuracy is shown in the bottom-right plot of Figure \ref{fig:hyperparams-n-interior}. Once again, we can see that very narrow networks obtain very poor accuracy, but the accuracy quickly converges as we keep adding neurons to the network. This quick saturation is again demonstrated in Figure \ref{fig:hyperparams-conv-prob1}. However, unlike the other three hyperparameters, we can see that the width of the network greatly impacts computation time. Whereas using only 16 neurons in each hidden layer yields nearly optimal accuracy in less than three seconds, increasing the width to 64 requires more than 18 seconds with little improvement in accuracy. This could be explained by the quadratic growth of the weight matrices between hidden layers, which increases both the computation time of evaluating the loss function as well as increasing the computation time of the L-BFGS vector product greatly. 

This last factor could be alleviated if we were to use a computationally cheaper optimizer. Specifically, the Adam optimizer and its derivatives are far more commonly used in neural network optimization, rather than second-order methods such as L-BFGS. To demonstrate the difference in convergence speed and final accuracy between these two optimizers, we plotted their NMAE as a function of time for \problemref{A}, see Figure \ref{fig:hyperparams-optimizer}. Note that, while Adam is traditionally used for stochastic optimization in machine learning, we are still using the same deterministic loss function as before in this example. As we can see in this figure, despite the increased computational and memory cost of L-BFGS, it converges significantly more quickly and achieves an accuracy that is several orders of magnitude better, suggesting second-order methods are still preferable for problems such as these. 

\section{Discussion}
\label{sec:discussion}
The results presented here indicate that neural network-based approaches can rival traditional methods in solving nonlinear PDEs. The primary advantage of these approaches lies in their simplicity; adapting the method to a specific PDE requires only the specification of a loss function, often achievable in just a few lines of code. The other components are largely generic, with the neural network resembling those used in various non-PDE applications and the L-BFGS solver being widely applicable across diverse fields, including but not limited to machine learning.

In contrast, the LSS is tailored specifically for this problem. Each step of its algorithm is meticulously designed to minimize the functionals described in Section \ref{sec:background-lss}. While this bespoke approach can potentially exploit the problem's unique characteristics, it demands greater expertise to implement and fine-tune. Notably, our results suggest that employing a general-purpose optimizer can yield accuracy and speed comparable to the LSS's specialized optimization techniques, and even outperform it in simple cases.

Furthermore, the current LSS implementation relies on a grid for computing the finite-differences-based Jacobian. This constraint necessitates the use of polar coordinates instead of Cartesian coordinates when dealing with circular source domains. Such a requirement not only introduces additional implementation complexity, but also potentially leads to slight artifacts, such as visible for \problemref{C}. In contrast, the MLP-L-BFGS solver circumvents this limitation, as it does not require interior points to be confined to a grid structure. This flexibility allows for more versatile problem-solving across various domain shapes without compromising accuracy or efficiency. 

Although our current approach employs deterministic optimization with a fixed set of interior and boundary points, it could be extended to incorporate stochastic optimization techniques. This modification would involve sampling a smaller subset of points at each iteration, significantly reducing the computational cost per iteration and potentially accelerating overall convergence time. Moreover, by resampling new interior points in each iteration, we may mitigate the risk of overfitting to specific points in the source domain, potentially yielding a more uniform distribution of error across the domain than observed in the results presented in Section \ref{sec:results}. However, such an adaptation would necessitate modifications to the L-BFGS algorithm to accommodate stochasticity, such as the approach proposed by \citet{moritz2016linearly}.

Nevertheless, our approach is not without challenges. The L-BFGS optimizer occasionally fails early, resulting in poor accuracy. Although this can be mitigated by reverting to first-order optimizers such as Adam or Rprop, doing so may impede convergence speed, even if adequate accuracy is eventually achieved. An alternative solution might involve learning an optimizer, a technique known as Learning to Optimize (L2O)\cite{li2016learning, chen2022learning, liao2023learning}. This could yield a more robust optimization routine that leverages the problem's specific structure without requiring manual optimizer tuning.

Moreover, while our approach performs adequately on the presented problems, its generalizability to other scenarios remains uncertain. Of particular concern are problems where the exact reflector contains a large number of high-frequency components, such as the `Girl with a Pearl Earring' problem examined by \citet{prins2015least}. Neural networks often struggle to represent such functions, especially when using common activation functions such as ReLU \cite{rahaman2019spectral}. Addressing these issues may necessitate the use of techniques such as random Fourier features \cite{tancik2020fourier}, sinusoidal activation functions \cite{rahimi2007random}, or alternative activation functions \cite{ramasinghe2022beyond}. 
% (???, wilbert: "?")

To better assess the performance and robustness of our method, it would be valuable to construct a larger, more representative test set of problems. While we have presented only five problems here, a much larger test set of representative problems could provide a more robust empirical estimate. Additionally, such a dataset could facilitate Neural Architecture Search (NAS)\cite{liu2021survey} to identify superior or more universally applicable architectures for complex reflector problems. \citet{wang2024pinn} explored a similar approach, though not for the Monge-Ampère equation, and their investigation was limited to the number and size of hidden layers. Future research can extend this approach to reflector problems and expand the architectural search space, potentially incorporating activation function selection through symbolic regression techniques\cite{orzechowski2018we}.

Finally, while we have only applied our method to parallel-to-far-field problems here, a number of other optical systems can be similarly characterized as solutions of the Monge-Amp\`ere equation with the transport boundary condition. As such, our approach could be extended to this broader class of problems. Future research could thus explore the application of our method to these problems to further assess its performance and generalizability.

\section{Conclusions}
\label{sec:conclusion}
This paper introduces a novel neural network-based approach to solving the Monge-Ampère equation with the transport boundary condition. This equation is crucial in various applications, including the design of reflectors to transform a given light source distribution into a desired far-field intensity pattern. Our method employs a carefully crafted loss function that incorporates the PDE residual, boundary condition, and convexity constraints, enabling the training of neural networks to approximate the equation's solution for multiple numerical examples. To our knowledge, this is the first paper to incorporate this unique boundary condition in a neural network-based PDE solution framework. 

Our proposed method demonstrated superior performance compared to a traditional least-squares finite difference solver across the test problems examined. A significant advantage of the neural network approach lies in its simplicity and versatility---the same fundamental methodology can be applied to a wide range of differential equations by modifying the loss function, eliminating the need for custom-derived update rules or optimization algorithms.

Despite these promising results, challenges persist in extending this approach to more complex problems. The sensitivity of the loss landscape occasionally complicates optimization, necessitating careful selection of optimizers, pre-training strategies, and architectures. Furthermore, the architecture and activation functions may limit the representation of high-frequency solution features. Future research could explore techniques such as neural architecture search and PDE-tailored learned optimizers to improve robustness and fidelity.

This work underscores the potential of neural networks as generalized PDE solvers when an appropriate loss formulation can capture the essential characteristics of the equation and solution space. By harnessing advances in deep learning methods and hardware, such approaches may offer a powerful and flexible alternative to traditional numerical PDE techniques. Continued research into stable architectures, training strategies, and comprehensive benchmarking will be vital to further establish neural networks as practical computational tools for solving differential equations across diverse scientific and engineering domains.

\clearpage
\bibliography{refs}

\end{document}